\pdfoutput=1
\documentclass[11pt]{article}

\usepackage[preprint]{acl}

\usepackage{times}
\usepackage{latexsym}

\usepackage[T1]{fontenc}
\usepackage[utf8]{inputenc} % allow utf-8 input
\usepackage{hyperref}       % hyperlinks
\usepackage{url}            % simple URL typesetting
\usepackage{booktabs}       % professional-quality tables
\usepackage{amsfonts}       % blackboard math symbols
\usepackage{nicefrac}       % compact symbols for 1/2, etc.
\usepackage{microtype}      % micro typography
\usepackage{xcolor}         % colors
\usepackage{multicol}
\usepackage{multirow}
\usepackage{makecell}
\usepackage{diagbox}
\usepackage{float}
\usepackage{graphicx}
\usepackage{amsmath}
\usepackage{amssymb}
\usepackage[scr=boondoxo]{mathalpha}  % heavily sloped
\usepackage{bbm}
\usepackage{mathtools}
\usepackage{amsthm}
\usepackage{caption}
\usepackage{subfigure}
\usepackage{calc}
\usepackage{enumitem}
\newcommand{\explain}[2]{%
  \underbrace{#1}_{%
    \text{\footnotesize\raggedright #2}% % 正确：在\text{}中
  }%
}

\usepackage[capitalize,noabbrev]{cleveref}

\usepackage{inconsolata}

\usepackage{pifont}

\title{Learn to Memorize: Scalable Continual Learning in Semiparametric Models with Mixture-of-Neighbors Induction Memory}

\author{Guangyue Peng\textsuperscript{\rm $\heartsuit$}\thanks{This work was done during the author's internship at Microsoft Research Asia.},
Tao Ge\textsuperscript{\ding{171}}\thanks{Corresponding author},
Wen Luo\textsuperscript{\rm $\heartsuit$},
Wei Li\textsuperscript{\rm $\heartsuit$},
Houfeng Wang\textsuperscript{\rm $\heartsuit$\textdagger} \\
{\textsuperscript{\rm $\heartsuit$} State Key Laboratory for Multimedia Information Processing,} \\
{School of Computer Science,  Peking University} \\ 
{\textsuperscript{\ding{171}} Microsoft} \\
{\tt \{agy,wanghf\}@pku.edu.cn,taoge@microsoft.com} \\ 
{\tt llvvvv22222@gmail.com,weili22@stu.pku.edu.cn} }

\begin{document}
\maketitle
\begin{abstract}

Semiparametric language models (LMs) have shown promise in various Natural Language Processing (NLP) tasks. However, they utilize non-parametric memory as static storage, which lacks learning capability and remains disconnected from the internal information flow of the parametric models, limiting scalability and efficiency. Based on recent interpretability theories of LMs, we reconceptualize the non-parametric memory represented by $k$NN-LM as a learnable Mixture-of-Neighbors Induction Memory (MoNIM), which synergizes the induction capabilities of attention heads with the memorization strength of feed-forward networks (FFN). By integrating into the model's information flow, MoNIM functions as an FFN-like bypass layer within the Transformer architecture, enabling effective learning of new knowledge. Extensive experiments demonstrate that MoNIM is a retentive and scalable continual learner in both data- and model-wise, enhancing the scalability and continual learning performance of semiparametric LMs.\footnote{Code is publicly available at \url{https://github.com/viniferagy/MoNIM}.}

\end{abstract}

\section{Introduction}

Semiparametric language models (LMs) have drawn increasing attention \citep{guu2020retrieval, yogatama_adaptive_2021, ram_context_2023} for their photographic memorization capabilities and in-domain accuracy. Combining a parameterized neural model with an extensible non-parametric memory, they are skillful at aiding prediction with memorization.

However, current semiparametric LMs are unsuitable for our fast-changing world because of inefficient memory management strategies \citep{he_efficient_2021}. They typically record all training data in the static memory, and the search for useful information then relies on additional modules or tunable hyperparameters. A lack of learning ability hinders memory compression, a process crucial for efficient learning in LMs \citep{deletang_language_2024}, and separates the memory component from the information flow of the model. As a result, these models experience linear growth of memory usage and search time as data or model dimensions increase. This inefficiency becomes especially impractical for large language models (LLMs), which deal with huge volumes of training data.

In this paper, we deal with two questions about semiparametric LMs: \textit{Why they are powerful but inefficient} and \textit{can we build an efficient memory strategy}? Enlightened by research interpreting the learning abilities of LLMs \citep{elhage_mathematical_2021,geva_transformer_2021,olsson_context_2022}, we propose that the non-parametric memory, specifically $k$NN-LM \citep{khandelwal_generalization_2020} inherently possesses abilities akin to the induction heads in Multi-Headed Self-Attention (MHSA) layers \citep{olsson_context_2022}. Perfect memorization brings perfect local next-token induction and good performance, but deficiencies in global visions prevent the memory from efficient reasoning. 

\begin{figure*}[ht]
\centering  %居中
\subfigure[]{   %第一张子图
\begin{minipage}{0.47\linewidth}
\centering    %子图居中
\includegraphics[width=\textwidth]{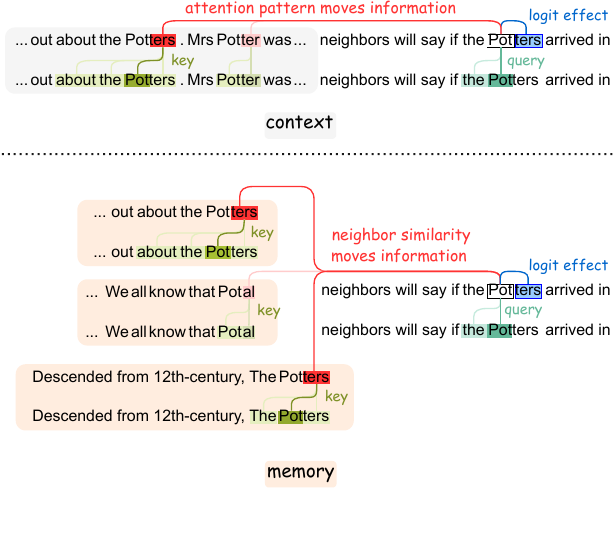}
\end{minipage}}
\subfigure[]{   %第一张子图
\begin{minipage}{0.49\linewidth}
\centering    %子图居中
\includegraphics[width=\textwidth]{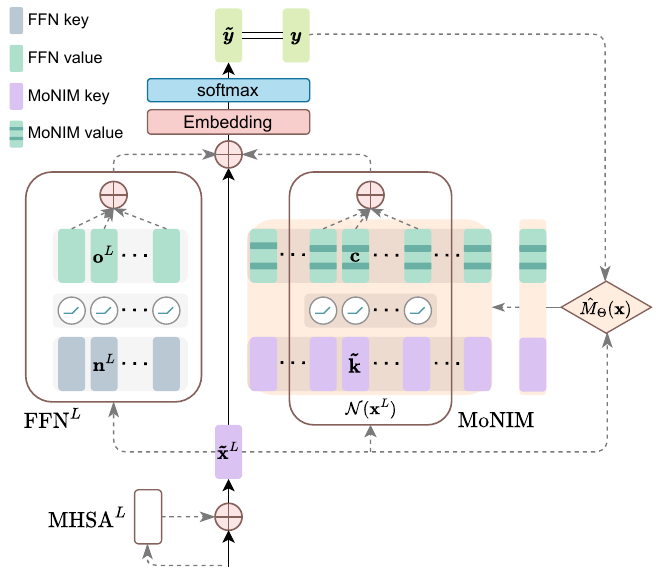}
\end{minipage}
}
\vspace{-0.3cm}
\caption{The learning mechanism of MoNIM. \textbf{(a)} The analogy between induction heads (up) and MoNIM (down). The shades of lines indicate allocated attention scores. While induction heads assimilate related information in context, MoNIM gathers similar samples memorized in learning history. \textbf{(b)} MoNIM as a learnable module. MoNIM shares the same input, working flow, and output with the final FFN layer. While FFN utilizes parametric keys $\boldsymbol{\mathrm{n}}^{L}$ to match the query $\tilde{\boldsymbol{\mathrm{x}}}^{L}$ for the promotion of learned concepts encoded in parametric values $\boldsymbol{\mathrm{o}}^{L}$ (green), MoNIM promotes concept mixtures $\boldsymbol{\mathrm{c}}$ (green with stripes) that are embedded in neighboring memorized values. The learning process of MoNIM is controlled by a compressor $\hat{M}_{\Theta}(\boldsymbol{\mathrm{x}})$ optimized for the model's loss function.}     %大图名称
\label{fig:model}    %图片引用标记
\end{figure*}

To promote local advantages while avoiding global weaknesses, we build a learnable Mixture-of-Neighbor Induction Memory (MoNIM) based on the components of concept promotion in Feed-Forward Networks (FFN) \citep{geva_transformer_2022}. As an FFN-like bypass layer, MoNIM can select and absorb knowledge with a gradually smaller memory footprint, demonstrating its consistency with gradient descent in parametric models, where the impact of new information is lessened with more data or larger model sizes \citep{kaplan_scaling_2020}. Consequently, MoNIM's memory grows sub-linearly with the enhancement of the model's capabilities, resulting in a scalable continual learner free of training. In continual learning scenarios, MoNIM performs comparably to vanilla models while consuming only half the memory space.

Our contributions can be summarized as follows:
\begin{itemize}[leftmargin=0.2in]
    \item We introduce Mixture-of-Neighbor Induction Memory (MoNIM), a learnable memory in semiparametric continual learning settings that functions as an FFN-like bypass layer. 
    \item MoNIM achieves its prowess by integrating the inductive capabilities of Multi-Head Self-Attention (MHSA) with the memorization functions of Feed-Forward Networks (FFN).
    \item Extensive experiments in language modeling and downstream tasks show that MoNIM effectively compresses seen information and is both data- and model-wise scalable, thus suitable for continual learning over streaming data with semiparametric LMs.
\end{itemize}

\section{MoNIM: Mixture-of-Neighbor Induction Memory}

\subsection{Preliminaries: $k$NN-LM}

Formally, we use $\Theta = (\theta, \mathcal{M})$  to denote a semiparametric LM, where $\theta$ stands for the parametric LM and $\mathcal{M}$ for the non-parametric memory.

As a representative, $k$NN-LM~\citep{khandelwal_generalization_2020} enhances the prediction of $\theta$ by leveraging the information of $k$-nearest neighbors in $\mathcal{M}$. Given a leftward context $\boldsymbol{\mathrm{x}}=(x_1,\dots,x_t)$, $k$NN-LM uses the hidden states in the final position before an FFN layer $\mathscr{l} \in \{1, .., L\}$ as the contextualized representation $\widetilde{\boldsymbol{\mathrm{x}}}^{\mathscr{l}} \in \mathbb{R}^d$, and computes its next word $y$'s probability as follows:
\begin{equation}\label{eq:slm}
P(y|\boldsymbol{\mathrm{x}};\Theta) = f(\explain{P(y|\boldsymbol{\mathrm{x}};\theta)}{\textrm{Model}}, \explain{P(y|\widetilde{\boldsymbol{\mathrm{x}}}^{\mathscr{l}};\mathcal{M})}{\textrm{Memory}}, \lambda)
\end{equation}
where $f$ represents the interpolation function to weigh the predictions of the model and memory by $\lambda$. $k$NN-LMs construct $\mathcal{M}$ by the training set $\mathcal{D}$ as a key-value lookup, with an entry for each token in the training set $x_t$ (as value) and the representation of its context $\widetilde{\boldsymbol{\mathrm{x_{<t}}}}^{\mathscr{l}}$ (as key):
\begin{equation}\nonumber
    \mathcal{M} = \{(\widetilde{\boldsymbol{\mathrm{x_{<t}}}}^{\mathscr{l}} \to x_t) | \mathrm{x} \in \mathcal{D} \}
\end{equation}

During inference, we first use $\tilde{\boldsymbol{\mathrm{x}}}^{\mathscr{l}}$ as a query to retrieve $k$ nearest neighbors from the memory $\mathcal{M}$:
\begin{equation}\nonumber
\mathcal{N}(\tilde{\boldsymbol{\mathrm{x}}}^{\mathscr{l}}) = \{ (\tilde{\boldsymbol{\mathrm{k}}} \to \tilde{y})_i | i=1, 2, \dots, k \} \subseteq \mathcal{M}
\end{equation}
Then, we obtain the prediction from $\mathcal{M}$ by computing the weighted sum of retrieved targets:
\begin{equation}\nonumber
P(y|\tilde{\boldsymbol{\mathrm{x}}}^{\mathscr{l}};\mathcal{M}) \propto \sum_{\mathcal{N}(\tilde{\boldsymbol{\mathrm{x}}}^{\mathscr{l}})} \mathbbm{1}_{\tilde{y}=y} \exp(-d(\tilde{\boldsymbol{\mathrm{k}}}, \tilde{\boldsymbol{\mathrm{x}}}^{\mathscr{l}})) 
\vspace{-0.2cm}
\end{equation}
here $d(.,.)$ denotes the semantic similarity. \citet{khandelwal_nearest_2021} proved that $\mathscr{l}=L$ produces the best retrieval quality.

\subsection{Induction Memory}

Recent interpretability studies \citep{olsson_context_2022,wang_interpretability_2023} have shown that induction heads, namely attention heads that implement a simple algorithm to complete sequences in the form of [A][B] ... [A] -> [B], might constitute the fundamental abilities for in-context learning in LLMs. Attention heads exhibit two typical properties: prefix matching, to attend to the tokens with similar context; and copying, to increase the logit of the output corresponding to the attended-to tokens \citep{bansal_rethinking_2023}. 

\cref{fig:model}(a) demonstrates the reasoning mechanisms between induction heads and the $k$NN memory. It is clear to observe their closeness: as induction heads assimilate related information scattered in the contexts, $k$NN memory gathers similar information from $\mathcal{N}(\tilde{\boldsymbol{\mathrm{x}}})$ memorized in history. The memory encodes the prefix into the key vector, uses it to match the query, and copies the memorized value to the position to be predicted as the function of induction heads. From this perspective, $k$NN memory is an induction buffer, considering all training data as neighbors, thus the name Mixture-of-Neighbor Induction Memory (MoNIM).

While induction heads serve as basic components in the early layers, transferring context information in the model to implement more complex global reasoning pathways, MoNIM, as a highly localized source of information, only provides the memorized labels. The capability to induce the most similar tokens also suggests its inability to perform efficient prediction. To make the best use of MoNIM's local precision, we move on to develop a new conceptual architecture of MoNIM.

\subsection{MoNIM for Local Concept Mixture Promotion}

According to previous work \citep{sukhbaatar_end_2015,geva_transformer_2021}, FFN layers function as key-value memory, and the value vectors can be projected to vocabulary space to represent comprehensible concepts such as food or movie characters \citep{geva_transformer_2022}. An FFN update $\mathcal{F}(\tilde{\boldsymbol{\mathrm{x}}})$ thus can be interpreted as successive concept promotions towards the connotation of the target token:
\begin{equation}\nonumber
\mathcal{F}^{\mathscr{l}}(\tilde{\boldsymbol{\mathrm{x}}}^{\mathscr{l}})=\sum_{i=1}^{d_\mathcal{F}}f(\tilde{\boldsymbol{\mathrm{x}}}^{\mathscr{l}}\cdot \boldsymbol{\mathrm{n}}^{\mathscr{l}}_i)\boldsymbol{\mathrm{o}}^{\mathscr{l}}_i=\sum_{i=1}^{d_\mathcal{F}}m^{\mathscr{l}}_i\boldsymbol{\mathrm{o}}^{\mathscr{l}}_i
\end{equation}

$\boldsymbol{\mathrm{n}}^{\mathscr{l}}_i, \boldsymbol{\mathrm{o}}^{\mathscr{l}}_i$ is the $i$-th column of $W_{in}^{\mathrm{T}}, W_{out}  \in \mathbb{R}^{d_{\mathcal{F}}\times d}$ in $\mathcal{F}$, $m^{\mathscr{l}}_i$ represents the weight of $\boldsymbol{\mathrm{o}}^{\mathscr{l}}_i$, where the model stores its understanding of concepts. Similarly, for the hidden state before an MHSA layer $\hat{\boldsymbol{\mathrm{x}}}$, given the attention pattern $\boldsymbol{\mathrm{a}}^{\mathscr{l}} \in \mathbb{R}^{T}$ for a context of length $T$ and corresponding $\boldsymbol{\mathrm{v}}^{\mathscr{l}}_j$, the j-th column of $W_{\mathrm{V}} \in \mathbb{R}^{T\times d_{\mathcal{A}}}$, an MHSA update $\mathcal{A}(\hat{\boldsymbol{\mathrm{x}}})$ is the linear combination of vectors of the output matrix $W_{\mathrm{O}} \in \mathbb{R}^{d_{\mathcal{A}}\times d}$.
\begin{equation}\nonumber
\mathcal{A}^{\mathscr{l}}(\hat{\boldsymbol{\mathrm{x}}}^{\mathscr{l}})=\sum_{j=1}^{d_\mathcal{A}}(\boldsymbol{\mathrm{a}}^{\mathscr{l}}\cdot \boldsymbol{\mathrm{v}}^{\mathscr{l}}_j)\boldsymbol{\mathrm{o}}^{\mathscr{l}}_j=\sum_{i=j}^{d_\mathcal{A}}m^{\mathscr{l}}_j\boldsymbol{\mathrm{o}}^{\mathscr{l}}_j
\end{equation}

Since \citet{geva_transformer_2022} has proved that the layer normalization (LN) is almost linear and does not affect the linear properties of MHSA and FFN outputs, we assert that the final prediction before the unembedding matrix $E \in \mathbb{R}^{|\mathcal{V}|\times d}$ can be decomposed to the reweighted sum of information gained in MHSA and FFN layers, that
\begin{equation}\nonumber
\tilde{\boldsymbol{\mathrm{x}}}^{\mathscr{l}}=\sum_{i=1}^{\mathscr{l}\cdot(d_{\mathcal{F}}+d_{\mathcal{A}})}\tilde{m}^{<\mathscr{l}}_i\boldsymbol{\mathrm{o}}^{<\mathscr{l}}_i
\end{equation}
\vspace{-0.3cm}
\begin{equation}\nonumber
\tilde{\boldsymbol{y}}=\mathrm{softmax}(E\tilde{\boldsymbol{\mathrm{x}}}^{L})
\end{equation}

The prediction of $\tilde{\boldsymbol{y}}$ is determined by the mixture of concepts $\boldsymbol{o}^{<L}$ in FFN. We define the best \textbf{local mixture of concepts} $\boldsymbol{\mathrm{c}}$ which outputs the golden prediction $\boldsymbol{y^{*}}$:
\begin{equation}\nonumber
\boldsymbol{\mathrm{c}}=\sum_{i=1}^{L\cdot(d_{\mathcal{F}}+d_{\mathcal{A}})}\tilde{m}^{*<L}_i\boldsymbol{\mathrm{o}}^{<L}_i
\end{equation}
\vspace{-0.3cm}
\begin{equation}\nonumber
\boldsymbol{y^{*}}=\mathrm{softmax}(E\boldsymbol{\mathrm{c}})
\end{equation}

We can infer that in MoNIM $\mathcal{M}$, the functioning form of memory entries is $(\tilde{\boldsymbol{\mathrm{k}}}\to \boldsymbol{\mathrm{c}})$, while the actual memory entries $(\tilde{\boldsymbol{\mathrm{k}}}\to \tilde{y})$ can be explained as economical and practicable appearance. MoNIM update can be viewed as a collection of sub-updates, each corresponding to a local mixture of concepts in the MoNIM output:
\begin{equation}\nonumber
\begin{aligned}
\mathcal{M}(\tilde{\boldsymbol{\mathrm{x}}}^{L})
=\sum_{\mathcal{N}(\tilde{\boldsymbol{\mathrm{x}}}^{L})}f_{\mathcal{M}}(\tilde{\boldsymbol{\mathrm{x}}}^{L}\cdot \tilde{\boldsymbol{\mathrm{k}}})\cdot\boldsymbol{\mathrm{c}}
=\sum_{\mathcal{N}(\tilde{\boldsymbol{\mathrm{x}}}^{L})}\tilde{m}\boldsymbol{\mathrm{c}}
\end{aligned}
\end{equation}

$(\tilde{\boldsymbol{\mathrm{k}}}\to \tilde{y})$ integrates into the information stream of Transformer for sake of $E$ to transform to $(\tilde{\boldsymbol{\mathrm{k}}}\to \boldsymbol{\mathrm{c}})$ to operate. Through $E$, MoNIM transforms into the general form as in $\mathrm{Eq}$ \ref{eq:slm}.
\begin{equation}\nonumber
\small
\begin{aligned}
\mathrm{log}P(y|\boldsymbol{\mathrm{x}};\Theta)&=\mathrm{log}\frac{\mathrm{exp}((1-\lambda)\boldsymbol{e}_{y}\boldsymbol{\mathrm{o}}^{L}+\lambda\boldsymbol{e}_{y}\mathcal{M}(\tilde{\boldsymbol{\mathrm{x}}}^{L}))}{Z(E((1-\lambda)\boldsymbol{\mathrm{o}}^{L}+\lambda\mathcal{M}(\tilde{\boldsymbol{\mathrm{x}}}^{L})))}\\
\propto &\mathrm{log}\frac{\mathrm{exp}((1-\lambda)\boldsymbol{e}_{y}\boldsymbol{\mathrm{o}}^{L})}{Z(E\boldsymbol{\mathrm{o}}^{L})}
\frac{\mathrm{exp}(\lambda\boldsymbol{e}_{y}\mathcal{M}(\tilde{\boldsymbol{\mathrm{x}}}^{L}))}{Z(E\mathcal{M}(\tilde{\boldsymbol{\mathrm{x}}}^{L}))}\\
= &(1-
\lambda)\mathrm{log}P(y|\boldsymbol{\mathrm{x}};\theta) + \lambda \mathrm{log}P(y|\widetilde{\boldsymbol{\mathrm{x}}};\mathcal{M})
\end{aligned}
\end{equation}
where $\boldsymbol{e}_{y}$ is the embedding of $y$, and $Z(\cdot)$ is the constant softmax normalization factor. \cref{fig:model}(b) demonstrates the equivalent working flows between $(\tilde{\boldsymbol{\mathrm{k}}}\to \boldsymbol{\mathrm{c}})$ in MoNIM and $(\boldsymbol{\mathrm{n}}^{\mathscr{l}}, \boldsymbol{\mathrm{o}}^{\mathscr{l}})$ in $\mathcal{F}^{L}$, formalizing MoNIM as an FFN-like bypass layer. From this perspective, MoNIM focuses on promoting local mixtures of concepts induced by the memorized neighbors, augmenting the induction abilities of LMs in the final layers. The complicated reasoning tasks are left to the parametric model which handles them better.

\section{MoNIM is a Scalable Continual Learner}

MoNIM's blend of memorization and induction suggests its potential to adapt to new knowledge, namely it can \textit{learn} as induction heads and compress worthless data for its induction task \citep{jiang2024longllmlingua,deletang_language_2024}. When the model confidently relies on global reasoning to tackle problems, MoNIM should step back to avoid impacting the model's performance. However, when the model lacks information for a decision, MoNIM should step in, promoting memorized local concept mixtures to help the model generate a more probable prediction.

\subsection{Learning Strategies of MoNIM}

We propose learning strategies for MoNIM that adopt cross-entropy, the optimization objectives of gradient descent.\footnote{Other possible strategies are discussed in \cref{appendix:learn_metric}.}

In gradient descent, the greater the cross entropy, the greater the gradient and the impact of data on parameter updates. For MoNIM learning, we indicate the same effect of data on memory capacity. The memory effect of a sample $M_{\Theta}(\boldsymbol{\mathrm{x}})$ can be expressed by its loss on model $\Theta$:
\begin{equation}\nonumber
    M_{\Theta}(\boldsymbol{\mathrm{x}}) \propto \log P(x_t|\boldsymbol{\mathrm{x_{<t}}};\Theta)
\end{equation}

To compress the data, rather than assign weights to indicate the importance of data points, we transform the weighted update into a "full-or-none" compressor $\hat{M}_{\Theta}(\boldsymbol{\mathrm{x}})$, namely only updates that weigh above a threshold $\delta$ will be saved into memory. Through this approximate method, we compress the unimportant part of data to take up no space and prove that the compressed part of data has very little effect on results.
\begin{equation}\label{eq:threshold}
    \hat{M}_{\Theta}(\boldsymbol{\mathrm{x}})=
    \begin{cases}
        1 & \text{if} ~ \log P(x_t|\boldsymbol{\mathrm{x_{<t}}};\Theta) < \delta \\
        0 & \text{else}
    \end{cases}
\end{equation}
\begin{equation}\label{eq:add}
    \mathcal{M} \gets \mathcal{M}~\cup~\{ \mathbbm{1}_{\hat{M}_{\Theta}(\boldsymbol{\mathrm{x}})}(\boldsymbol{\widetilde{\mathrm{x_{<t}}}} \to x_t)\}
\end{equation}

\subsection{Adaptive MoNIM weight}
\label{sec:ada}

Instead of using a fixed threshold ($\delta$ in $\mathrm{Eq}$ \ref{eq:threshold}), we propose to use an adaptive memorization threshold (AMT) to enhance the effect of MoNIM:\footnote{The ablation of AMT is placed in \cref{appendix:amt}.}

\begin{equation}\nonumber
    \delta_{\text{ada}}=
    \frac{\delta}
    {\max_{y}
    \log\frac
    {P\left(y|\boldsymbol{\mathrm{x_{<t}}};\Theta\right)}
    {P\left(x_{t}|\boldsymbol{\mathrm{x_{<t}}};\Theta\right)}
    +0.5
    }
\end{equation}

The best form of threshold is not the focus of this paper, however, we found that AMT-like types of threshold boost the experimental results. The intuition of AMT is straightforward: if $\max _{y} \log P\left(y|\boldsymbol{\mathrm{x}}_{<\boldsymbol{t}}; \Theta\right)==\log P\left(x_{t}|\boldsymbol{\mathrm{x}}_{<\boldsymbol{t}}; \Theta\right)$, then the memorization margin $\delta_{\text {ada}} \leftarrow 2 \delta$ ( $\delta$ is the base threshold), meaning we can relax the threshold to $2 \delta$ since $x_{t}$ is already the top-1 prediction and thus not urgent to be memorized. On the contrary, if $\max _{y} \log P\left(y|\boldsymbol{\mathrm{x}}_{<\boldsymbol{t}}; \Theta\right)>>\log P\left(x_{t}|\boldsymbol{\mathrm{x}}_{<\boldsymbol{t}}; \Theta\right)$, then $\delta_{\text{ada}}<\delta$, indicating we should aggressively memorize this sample because of the large gap between it and the top-1 prediction.

AMT is simple yet effective in practice. It allows us to skip many samples with the top rank, substantially reducing the memory size with marginal generation quality loss; moreover, it alleviates overfitting top-ranked samples, playing a similar role as label smoothing to avoid overconfident predictions.

Since MoNIM updates, unlike FFN layers \citep{geva_transformer_2022}, always promote concepts rather than eliminate or run shortcuts, if extracted neighbors are so thin and scattered that there is no reliable concept to promote, its weight $\lambda$ should be pushed down accordingly. Inspired by previous studies \citep{he_efficient_2021,drozdov_you_2022}, we train a simple calibrator to inform the reliability of MoNIM with three categories of features: distribution information of the parametric LM, lexical information of the training data, and density information of MoNIM. Following \citet{he_efficient_2021}, we use a 4-layer MLP network, optimized on a small subset of the validation set.\footnote{The detailed implementation of the calibrator is placed in \cref{appendix:cali_details}, and the ablation study in \cref{appendix:cali_abl}.}

\subsection{Scalability of MoNIM}
% \vspace{-0.2cm}

Unlike the space-inefficient $k$NN-LM, MoNIM's learning capability allows it to compress and reduce memory demand throughout the learning process. We designed experiments in continual learning settings showing that MoNIM can keep compressing when updating. Further, we reveal that the features of compression are consistent with those of updates of model parameters and lead MoNIM to scalability: \textbf{(i) data scalability}: In parametric models, as learning progresses, the impact of data on parameter updates diminishes; similarly, in MoNIM, as it continues to learn, the influence of data on memory capacity diminishes, meaning less new information needs to be memorized. \textbf{(ii) model scalability}: As parametric models grow in size, the impact of data on parameter updates decreases; likewise, in MoNIM, the impact of data on memory capacity also diminishes with model growth.

\section{Experiments}

\subsection{Experimental Setting}

We use the news from December 2019 in the News Crawl corpus (NC-19Dec) as pilot data, and apply \textbf{Newscrawl-20H1} (NC-20H1), the articles during the first half of 2020 in the News Crawl, as our streaming data for continual learning (CL). We randomly select 1\% data per day as the validation and test set, and the rest 98\% articles as the training set. \cref{table:news_statistics} shows the statistics of NC-20H1. We continually learn the streaming data in chronological order and update the search index\footnote{Implementation of search is included in \cref{appendix:app_index}.} every day. 

In addition, we construct \textbf{WikiEvent-20H1} (WE-20H1), a Wikipedia event dataset\footnote{An example event article is \href{https://en.wikipedia.org/wiki/2020_Caribbean_earthquake}{2020 Caribbean earthquake}.} describing real-world events during 20H1, for testing our approach in domains other than news. WE-20H1 contains, on average 10 Wikipedia articles per month with $\sim$100k tokens in total.

We use GPT-2~\citep{radford_language_2019} as the backbone LM to study CL over 20H1's streaming data. We experiment with the GPT-2 small (S, 123M), medium (M, 355M), and large (L, 774M) variants\footnote{The detailed configurations are in \cref{appendix:model_config}.}, and GPT-2 small is assumed to be the default size unless otherwise specified. All the experiments are implemented using the Fairseq (\citealp{ott_fairseq_2019}) toolkit and run on 1 NVIDIA V100 GPU.

We define the memorization rate (MemRate) as the percentage of key-value pairs stored in memory compared to the \textit{FullMem} baseline. MemRate is utilized to measure the memory efficiency and scalability of our method. It comes that for NC-19Dec, when $\delta=-1.5$, MoNIM can achieve comparable performance to \textit{FullMem} with $\sim$60\% MemRate. Thus, we set $\delta=-1.5$\footnote{We explored the effect of different choices of $\delta$ on performance and memorization in \cref{appendix:memrate_abl}.} throughout our following experiments.

After CL, we conducted extensive experiments in both language modeling and downstream tasks to estimate MoNIM's performance and scalability. MoNIM is compared with the following baselines:
\begin{itemize}[leftmargin=0.2in]
\item Full memorization (\textit{FullMem}): Conventional memorization policy that memorizes every token in the training set.
\item Random memorization (\textit{RandMem}): Randomly memorize data equal to MoNIM's initial MemRate (60\%). We conduct three runs with random seeds and choose the best as the baseline.
\end{itemize}

\begin{table}[htbp]
\centering
\resizebox{\linewidth}{!}{
\begin{tabular}{c|cc|cc|cc}
\toprule
     &\multicolumn{2}{c}{\textbf{Daily}}&\multicolumn{2}{|c|}{\textbf{Monthly}}&\multicolumn{2}{c}{\textbf{Total}}       \\
     &\#Train&\#Dev/Test &\#Train&\#Dev/Test &\#Train&\#Dev/Test \\
\midrule
Articles     &4.4K&46       &133K&1.3K &796K&8.2K             \\
Tokens       &2.4M&24.7K &73.2M&741K      &439M&4.5M            \\
\bottomrule
\end{tabular}
}
\caption{Statistics of NC-20H1.}
\label{table:news_statistics}
\end{table}

\begin{figure*}[ht]
\centering
\includegraphics[width=0.85\textwidth]{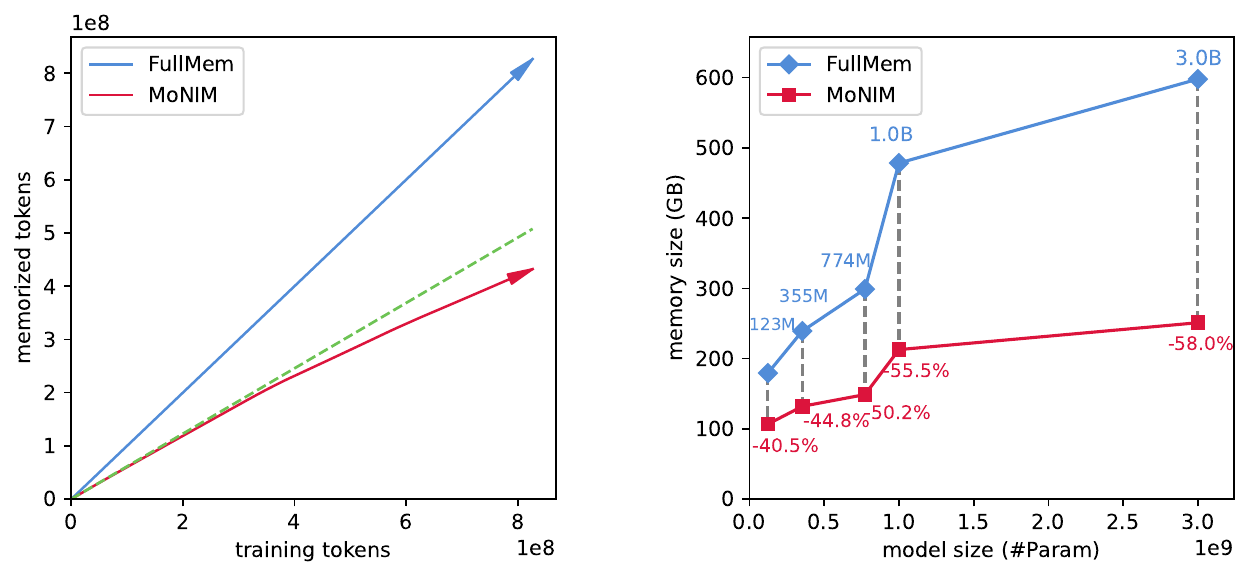}
\vspace{-0.2cm}
\caption{\textbf{(a)} The memory growth of the conventional full memorization (\textit{FullMem}) policy and our proposed MoNIM policy, whose performance is controlled to be comparable with that of \textit{FullMem}. Not only does MoNIM significantly reduce its memory but also the growth of its memory will become slower than when it starts (i.e., the green dashed line), as it continuously learns; \textbf{(b)} The growth trend of memory size (GB) with the increase of the model size (from 123M to 3B). The red numbers below the MoNIM's line indicate the memory consumption reduced by MoNIM, showing that MoNIM's effect is more remarkable in larger LMs.}
\label{fig:intro}
\vspace{-0.2cm}
\end{figure*}

\subsection{Results of language modeling}

\subsubsection{Data scalability}

\cref{table:lm_compare_strategy} compares between MoNIM and \textit{FullMem}, \textit{RandMem} for CL with NC-20H1, showing substantial improvements for the bare GPT-2 small. Among them, MoNIM achieves comparable (even slightly better) language modeling performance to \textit{FullMem} but with only 50\% MemRate, and largely outperforms \textit{RandMem} (60\% MemRate), demonstrating that MoNIM is a cost-effective memorization policy for CL.

\begin{table}[ht]
\centering
\resizebox{0.9\columnwidth}{!}{
\begin{tabular}{l|c|c}
\toprule
Methods     & PPL ($\downarrow$) & MemRate ($\downarrow$) \\
\midrule
Bare GPT-2 small           & 24.1         & 0\%              \\
+\textit{FullMem}      & 9.0          & 100\%            \\
+\textit{RandMem}      & 15.0         & 60\%             \\ \midrule
+MoNIM  & \textbf{8.6} & \textbf{50\%}    \\
\bottomrule
\end{tabular}
}
\caption{Perplexity (PPL) on the test set of NC-20H1 by different memorization methods for CL over NC-20H1.}
\label{table:lm_compare_strategy}
\end{table}

We look into the results by tracking MoNIM's monthly memorization throughout CL. The sublinear growth trend of MoNIM observed in Figure \ref{fig:intro}(a) indicates that the compression rate of MoNIM gradually increases as it learns over time because the model becomes increasingly knowledgeable and skips more training cases that it already knows. 

\begin{table}[t]
\centering
\resizebox{\columnwidth}{!}{
\begin{tabular}{c|c|c|c|c}
\toprule
\multirow{2}{*}{\textbf{Model}} & \multicolumn{2}{c|}{\textbf{\makecell{NC-20July \\ \small{(same domain)}}}} & \multicolumn{2}{c}{\textbf{\makecell{WE-20H1 \\ \small{(different domain)}}}} \\ 
                   & MemRate     & PPL   & MemRate     & PPL            \\ \midrule
GPT-2 small               &       60\%                        &    8.8     &          62\%        &    29.0    \\
+CL w/ NC-20H1      &     \bf   44\%                &     \bf    8.2   &          \bf     57\%         &   \bf  27.5  \\ \bottomrule 
\end{tabular}
}
\caption{Comparison of memorization rate and perplexity for new CL data -- NC-20July and WE-20H1 before and after CL with NC-20H1.}
\label{tab:data-wise}
\end{table}

We further confirm MoNIM's data-wise scalability by comparing the MemRates of two additional CL datasets — news data from July 2020 in the News Crawl (NC-20July) and WE-20H1 — before and after CL with NC-20H1. Following CL on NC-20H1, we utilize MoNIM (equipped with the memory acquired from NC-20H1) to continually process NC-20July and WE-20H1. As shown in Table \ref{tab:data-wise}, MoNIM's MemRates for both datasets are significantly lower than those of the models without CL, while improving performance. This reduction in MemRates can be attributed to the prior CL, which enables the model to avoid memorizing numerous instances learned previously, without compromising performance.

\subsubsection{Model scalability}

\cref{table:lm_model_size} shows the model size's effect on MoNIM. In the premise of comparable results to \textit{FullMem}, the MoNIM's effect becomes more significant as the parametric model becomes larger: its overall MemRate drops from 50\% (GPT-2 small) to 40\% (GPT-2 large). As we assumed, a larger LM tends to skip more training cases than a smaller LM. The reduced MemRate demonstrates the model-wise scalability of MoNIM.

\begin{table}[t]
\centering
\resizebox{0.9\columnwidth}{!}{
\begin{tabular}{c|c|c|c}
\toprule
Model      & Strategy  & PPL ($\downarrow$) & MemRate ($\downarrow$) \\ 
\midrule
\multirow{2}*{\makecell{S\small{(123M)}}}  & \textit{FullMem}        & 9.0     & 100\%              \\ \cline{2-4}
                           & MoNIM       & 8.6     & 50\%              \\
\hline
\multirow{2}*{\makecell{M\small{(355M)}}} & \textit{FullMem}        & 7.0     & 100\%              \\ \cline{2-4}
                           & MoNIM       & 7.2     & 46\%              \\
\hline
\multirow{2}*{\makecell{L\small{(774M)}}}  & \textit{FullMem}        & 6.2     & 100\%              \\ \cline{2-4}
                           & MoNIM       & 6.3     & 40\%              \\
\bottomrule
\end{tabular}
}
\caption{MoNIM's model-wise scalability for GPT-2 models of different sizes.}
\label{table:lm_model_size}
\end{table}

% 5/14 add %
To test the generalizability of scalable memory on larger LMs, we choose Meta's Llama-3.2-1B and 3B versions \citep{dubey2024llama} for a brief evaluation of MoNIM's performance\footnote{Since Llama-3.2 was released on September 25, 2024, we extract news from the first week of October 2024 in the News Crawl corpus to implement this experiment.}. As shown in \cref{fig:intro}(b), the total MemRate reduces from GPT2-small’s 59.5\% (123M) to Llama-3.2-3B’s 42.0\% (3B). We assume that larger LMs have the potential to achieve even more negligible memory consumption, as long as MoNIM maintains both data- and model-scalability.

\begin{table*}[ht]
\centering
\resizebox{0.85\textwidth}{!}{
\begin{tabular}{l|c|c|c|c|c|c|c}
\toprule
\textbf{Methods} & \textbf{Wiki-103} & \textbf{1 (Jan)} & \textbf{2 (Feb)} & \textbf{3 (Mar)} & \textbf{4 (Apr)} & \textbf{5 (May)} & \textbf{6 (Jun)} \\ \midrule
Bare GPT-2 & 29.1 & 24.3 & 24.0 & 24.1 & 24.0 & 24.7 & 23.8  \\ \midrule
+Fine-tune &   33.4 (+2.9)  & 20.4 (+2.4)       & 18.7 (+2.3)       & 17.3 (+1.3)       & 17.3 (+0.2)       & 16.2 (+0.8)       & 15.4 (+0.0)       \\
~~~~~(best)    &    30.6   & 18.2       & 16.4       & 16.0       & 17.1       & 15.4       & 15.4       \\ \midrule
+RecAdam   &    34.5 (+2.7)        & 19.6 (+0.7)      & 18.3 (+0.7)       & 17.3 (+0.2)       & 17.1  (+0.3)      & 16.8 (+0.3)      & 16.9 (+0.0)      \\
~~~~~(best)    &   31.8    & 18.9       & 17.6       & 17.1       & 16.8       & 16.5       & 16.9       \\ \midrule
+MixReview &    33.6 (+3.0)     & 19.9 (+1.8)      & 18.5 (+2.1)      & 17.3 (+1.4)       & 17.2  (+0.1)      & 15.8 (+0.4)      & 15.6 (+0.0)      \\
~~~~~(best)    &   30.6    & 18.1       & 16.4       & 15.9       & 17.1       & 15.4       & 15.6       \\ \midrule
+Greedy Merging &    35.2 (+5.9)     & 15.3 (+6.3)      & 15.5 (+5.2)      & 16.0 (+3.1)       & 15.9  (+2.8)      & 15.8 (+1.8)      & 14.6 (+0.0)      \\
~~~~~(best)    &   29.3    & 9.0       & 10.3       & 12.9       & 13.1       & 14.4       & 14.6       \\ \midrule
+MoNIM     &   \textbf{29.9} (+0.5)  &  \textbf{9.4} (+0.1)  &  \textbf{7.6} (+0.2)  &  \textbf{7.8} (+0.5)  &  \textbf{6.9} (+0.1)  &  \textbf{9.5} (+0.0)  &  \textbf{8.8} (+0.0)  \\
~~~~~(best)    &  29.4  &  9.3  &  7.4  &  7.3  &  6.8  &  9.5  &  8.8  \\ \bottomrule
\end{tabular}}
\caption{PPL evaluated on the 7 test sets after CL over NC-20H1 between MoNIM and representative CL approaches. The numbers in the second row of each cell denote the best result achieved during the process of CL.}
\label{tab:cl_baselines}
\end{table*}

\begin{figure}[t]
\centering
\includegraphics[width=\linewidth]{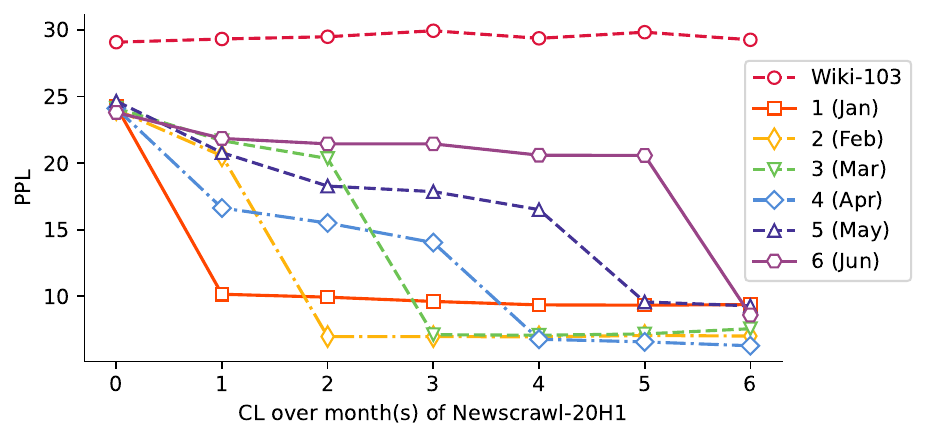}
\caption{Language modeling performance on 7 test sets (Wiki-103 and 6 subsets of Newcrawl-20H1's test set by month) throughout CL via MoNIM over NC-20H1.}
\label{fig:lm_ppl}
\end{figure}

\begin{table}[t]
\centering
\resizebox{\columnwidth}{!}{
\begin{tabular}{l|c|c|c} 
\toprule
\textbf{Model (MemRate)}  & \textbf{Wiki-103}       & \textbf{NC-20H1} & \textbf{ACL} \\ \midrule
Bare GPT-2 (0\%) & 29.1                    & 24.1                    & 40.5         \\ \midrule
+\textit{FullMem} (100\%)  & 30.1 $\to$ 31.0 & 9.0 $\to$ 10.0  & \textbf{22.5}         \\ \midrule
+MoNIM (64\%)     & 29.9 $\to$ \textbf{30.4} & 8.6 $\to$ \textbf{9.2}  & 22.7         \\ \bottomrule
\end{tabular}
}
\caption{CL performance over ACL papers after learning NC-20H1. The numbers beside each arrow indicate the PPL before/after studying the ACL papers.}
\label{tab:acl}
\end{table}

\subsubsection{Mitigation of forgetting}

MoNIM's learning against forgetting performance is evaluated by tracking results on 7 test sets throughout CL over the NC-20H1: 6 are each month's held-out data in NC-20H1, and the other is the test set of Wiki-103 benchmark (\citealp{merity_pointer_2017}) which does not overlap with the NC-20H1 training data. According to Figure \ref{fig:lm_ppl}, MoNIM learns from the streaming data well, reflected by the sharp decrease of PPL on a test set after learning its corresponding month's training data. More importantly, it does not suffer much from the catastrophic forgetting issue (\citealp{french_catastrophic_1999}). PPL scores of all 7 test sets do not significantly degrade throughout CL, since MoNIM will never erase previous memory or update the weights of LM.

The advantage can be better understood by comparing MoNIM with other CL baselines. We select two popular CL methods, RecAdam~\citep{chen_recall_2020} and Mix-Review~\citep{he2021analyzing}, with Greedy Merging, the most effective data compression approach in \citet{he_efficient_2021}.\footnote{Details of CL baselines are included in \cref{appendix:cl_baselines}.} As in Table \ref{tab:cl_baselines}, MoNIM not only achieves better results in learning from the new data but also suffers less from catastrophic forgetting than other CL approaches despite introducing additional memory.

\begin{figure}[t]
\centering
\includegraphics[width=\linewidth]{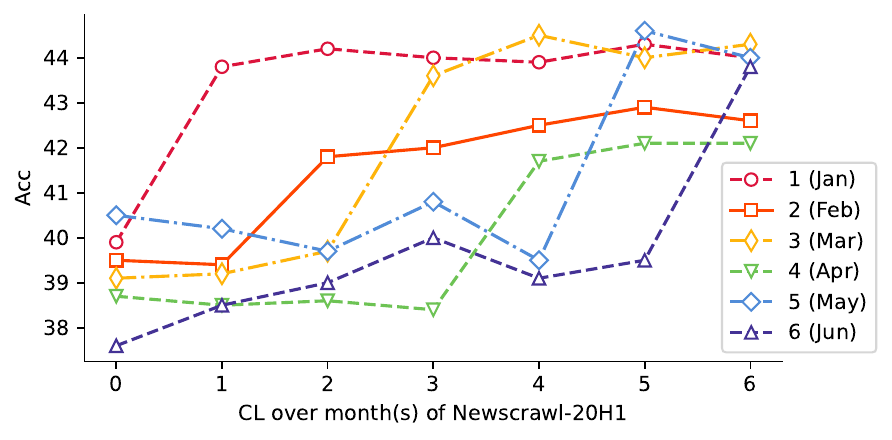}
\caption{Next-word prediction accuracy on 6 subsets (by month) of WE-20H1 throughout CL over NC-20H1.}
\label{fig:wiki_acc}
\end{figure}

\begin{table}[t]
\centering
\resizebox{\linewidth}{!}{
\begin{tabular}{l|cccc|ccc}
\toprule
\multirow{2}{*}{Methods}     & \multicolumn{3}{c}{Acc ($\uparrow$)}  &   & \multicolumn{3}{c}{MemRate ($\downarrow$)} \\\cline{2-5}\cline{6-8}
           & S & M & L && S & M & L \\ 
\midrule
Bare GPT-2         & 39.5  & 41.2  & 44.6   && 0\%   & 0\%    & 0\%    \\
+\textit{FullMem}    & 43.0  & \bf 45.5  & \bf 48.9   && 100\% & 100\%  & 100\%  \\
+\textit{RandMem}    & 40.3  & 41.3  & 45.5   && 60\%  & 60\%   & 60\%   \\ \midrule
+MoNIM & \textbf{43.8} & 45.4 & 48.5 && \textbf{50\%}  & \textbf{46\%} & \textbf{40\%}     \\
\bottomrule
\end{tabular}
}
\caption{Next-word prediction accuracy (Acc) on the test set (WE-20H1) and MemRate by different memorization methods and models after CL over NC-20H1.}
\label{table:wiki_size}
\end{table}

\subsubsection{Domain adaptation}

After CL over the news streaming data, which is not of great difference from the LM's pretraining dataset (i.e., openwebtext for GPT-2), we test MoNIM's CL performance over data in another domain -- the ACL paper dataset \citep{lo_s2orc_2020} with 42K ACL papers. We hold off 80 papers ($\sim$200K tokens) to construct the validation and test set, using the rest for training.\footnote{We split the training data into 4 batches for CL and update the index after finishing each batch.} \cref{tab:acl} shows that utilizing less memory (64\% compared with \textit{FullMem}), MoNIM consistently performs well in new data (40.5 $\to$ 22.7 in PPL) with less forgetting degradation. Although \textit{FullMem} is also relatively resilient to catastrophic forgetting, it stores more noise from in-domain samples, which can degrade retrieval performance when applied to out-of-domain inputs.

\subsection{Results of downstream tasks}

\subsubsection{Next-word prediction}

For LLM, next-word prediction is the basic and the most straightforward end task, especially important for AI applications (e.g., input methods, \href{https://insider.office.com/zh-cn/blog/text-predictions-in-word-outlook}{Microsoft's text predictions} and \href{https://chat.openai.com}{ChatGPT}).

To align this task with our CL setting, we test the next-word prediction on the WE-20H1 to verify if CL over NC-20H1 can help write the current event articles in Wikipedia. As in language modeling, MoNIM consistently shows comparable performance with better scalability than FullMem as the model size increases (Table \ref{table:wiki_size}) and desirable results with little forgetting (Figure \ref{fig:wiki_acc}).

\subsubsection{Closed-book question answer}

We use \textsc{RealTime QA} (\citealp{kasai2023realtime}), a multiple-choice question dataset about real-time events, as our second testbed of downstream tasks. To align with our streaming data, we use the subset of news during 20H1 and evaluate it in the closed-book Multiple Choice setting. As \citet{kasai2023realtime} suggests, we evaluate GPT-2 large in a zero-shot learning setting, in which GPT-2 small and medium are too weak to perform.

% to do: no multi-choice? no curve? use zero-shot?
\cref{table:qa_compare_strategy} shows the results in \textsc{RealTime QA}. Compared with the bare GPT-2 large, CL through MoNIM over NC-20H1 substantially improves QA performance because it learns the world knowledge during 20H1 from the news stream to answer the questions. MoNIM again performs as well as FullMem with less memory footprint and outperforms RandMem. Moreover, little forgetting is consistently observed, as shown in Table \ref{table:qa_acc}.

\begin{table}[t]
\centering
\resizebox{0.75\linewidth}{!}{
\begin{tabular}{l|c}
\toprule
Methods (MemRate)     & Acc ($\uparrow$)  \\
\midrule
Bare GPT-2 large (0\%)           & 29.8                    \\
+\textit{FullMem} (100\%) & \bf 36.3                    \\
+\textit{RandMem} (60\%)       & 30.7                     \\ \midrule
+MoNIM (\textbf{40\%}) & 36.2     \\
\bottomrule
\end{tabular}
}
\caption{Accuracy on \textsc{RealTime QA} of GPT-2 large with CL over NC-20H1 in zero-shot learning setting.}
\label{table:qa_compare_strategy}
\end{table}

\begin{table}[t]
\centering
\resizebox{0.8\linewidth}{!}{
\begin{tabular}{c|ccc}
\toprule
\diagbox{CL over}{Test on}  & 1-2 & 3-4 & 5-6  \\ 
\midrule
1-2      & 36.0  & 29.3  & 29.5    \\
1-4      & 37.2  & 35.8  & 30.6    \\
1-6      & 37.0  & 36.5  & 35.5   \\
\bottomrule
\end{tabular}
}
\caption{Accuracy on every two months of GPT-2 large of \textsc{RealTime QA} throughout CL over NC-20H1.}
\label{table:qa_acc}
\end{table}

\section{Related Work}

\paragraph{Retrieval-augmented LMs} 
Retrieval components have been found beneficial for language tasks. Unlike the explicit storage methods \citep{guu2020retrieval,borgeaud_improving_2022}, semiparametric LMs like $k$NN-LM (\citealp{khandelwal_generalization_2020}) store implicit information as key-value pairs to assist prediction, without the need for retraining. As a powerful method to use the external data, many successive works of $k$NN-LM have been proposed~\citep{zheng_adaptive_2021,jin_plug_2022,trotta_nearest_2022,bhardwaj-etal-2023-adaptation}. Among them, \citet{he_efficient_2021} focuses on improving the efficiency, which has similar applications as our work, but we focus more on interpretable scalability in CL over streaming data with orthogonal contributions. 

\paragraph{Interpretable LMs} 
It remains obscure how Transformer manages to understand and generate natural languages. Among all the struggles to open the black box, mechanistic interpretability \citep{elhage_mathematical_2021,olsson_context_2022,gurnee_finding_2023} investigates neurons and their connections in terms of circuits where information flows and transforms. Previous works have found many components that provide learning capabilities \citep{merullo_circuit_2024,zhang_towards_2024}. We transplant these interpretations that work for the parametric models to non-parametric memory, which has not been reasonably explained. Experiments prove the feasibility of our conceptual framework of MoNIM.

\paragraph{Continual learning LMs}
Continual learning (CL) proposes to address the “new knowledge - catastrophic forgetting” dilemma (\citealp{french_catastrophic_1999}). According to the taxology of \citet{wang_comprehensive_2024}, our method deals with catastrophic forgetting problems based on replay-based methods (\citealp{sun_lamol_2019,qin_elle_2022,scialom_fine_2022}), despite that we managed to build a learnable replay memory. CL for LM is gaining traction \citep{ke_continual_2022,razdaibiedina_progressive_2023}, and the closest works to us are \citet{jang_towards_2022} and \citet{jin_lifelong_2022}, which adapt LMs to emerging corpora across domains and timelines. However, we are the first to explore the memory to deal with non-parametric solutions for CL over streaming data.

\section{Conclusion}

We introduced Mixture-of-Neighbors Induction Memory (MoNIM), a novel conceptual framework that integrates dynamic induction memory into the Transformer architecture to interpret and enhance semiparametric LMs. Our experiments demonstrate that MoNIM not only offers a fresh perspective on non-parametric memory but also sets a new benchmark for scalable and efficient learning in LLMs, giving insights for the evolution of LLMs without the need for parameter adjustments.

\section*{Limitations}

We construct the framework of MoNIM and thoroughly investigate its practicality and effectiveness as a representative of semiparametric LMs in continual learning. However, MoNIM only formalizes the interpretation of $k$NN-LM, while there are diverse models and implementations under semiparametric LMs. Specifically, besides auto-regressive models, auto-encoder models like T5 (\citealp{jang_towards_2022}) also exhibit their potential for continual LM. Although we have observed that T5s are empirically capable of continual learning, the framework we constructed does not currently include them. In the future, we intend to extend to varied models and architectures and confirm the universal effectiveness of our framework.

Due to resource constraints, we tested our method on data within half a year (20H1) and on models up to 3B in size. Stretching the time series and increasing the model size is urgent for observing a more prominent and convincing build-up curve for a longer period.

\section*{Acknowledgments}

This work was supported by National Natural Science Foundation of China (62036001) and National Science and Technology Major Project (No. 2022ZD0116308) . The corresponding author is Houfeng Wang.

% Entries for the entire Anthology, followed by custom entries
\bibliography{paper-rebiber}

\appendix

\section{Experiment details}

\subsection{Model configuration}
\label{appendix:model_config}

We list the model configurations of the GPT-2 models (as the parametric LMs) in our experiments in \cref{tab:gpt2}.

\begin{table}[htbp]
\centering
\begin{tabular}{c|c|c|c}
\toprule
Model &  Layer & Dim & \#Param \\ \midrule
GPT-2 small & 12 & 768 & 123M \\ 
GPT-2 medium & 24 & 1024 & 355M \\
GPT-2 large & 36 & 1280 & 774M \\ \bottomrule
\end{tabular}
\caption{Model configurations of the GPT-2 models (as the parametric LMs) in our experiments.}
\label{tab:gpt2}
\end{table}

\subsection{Index building}
\label{appendix:app_index}
We use the FAISS toolkit \citep{johnson_billion_2019} for index building and searching. At each update, we sampled 1M keys randomly from memory to train 4K cluster centroids, and then the whole keys in memory are added to the trained index, all quantized to 64-bytes. 

\subsection{NN calibrator training}
\label{appendix:cali_details}

We list the features we use in training the NN calibrator:

\begin{itemize}
    \item Distribution information of the parametric LM
    \begin{itemize}
        \item $\tilde{\boldsymbol{\mathrm{x}}}$: contextualized representation of $\boldsymbol{\mathrm{x}}$ by the parameterized LM
        \item $conf(\boldsymbol{\mathrm{x}})$: $\max_y P(y|\boldsymbol{\mathrm{x}}; \theta)$
        \item $ent(\boldsymbol{\mathrm{x}})$: entropy of $P(y|\boldsymbol{\mathrm{x}}; \theta)$
    \end{itemize}
    \item Lexical information of the training data
    \begin{itemize}
        \item $\log freq(\boldsymbol{x_{-1}})$: log of frequency of the last token in the context
        \item $\log distinct(\boldsymbol{\mathrm{x_{-1}}})$: log of the number of distinct values that succeed the last token in the context
    \end{itemize}
    \item Density information of the external memory
    \begin{itemize}
        \item $d(\tilde{\boldsymbol{\mathrm{k}}}, \tilde{\boldsymbol{\mathrm{x}}})$: $L^2$ distance (semantic similarity) between the query and the top-$i$ retrieved neighbor, $i=1, 2, \dots, 10$.
        \item $\log distinct(\tilde{y})$: log of the number of distinct values of the top-$i$ retrieved values, $i=1, 2, \dots, 10$.
    \end{itemize}
\end{itemize}

On each day during memorization, we extract from the validation set 10 articles and update them into the training set of the calibrator yesterday to obtain the training set of the calibrator today. The validation set of the calibrator is obtained as above, except for it only needs 5 articles each day. Because the training set increases slowly every day, we reduce the number of training epochs from 5 epochs to 1 epoch as time goes on, in case of overfitting.

Each feature is fed into a 1-layer LeakyReLU network to be transformed into hidden states of 128-dimension equally. Then all the hidden states are concatenated to a long vector and fed into a 4-layer MLP network to predict $\lambda$. We list the hyperparameters of the NN calibrator in \cref{table:appendix_hyp}.

\begin{table}[htbp]
\centering
\begin{tabular}{lr}
\toprule
Hyperparameters        & Values \\
\midrule
Layers                 & 4      \\
Dimension of hidden state          & 128    \\
Learning rate          & 3e-4     \\
Optimizer              & Adam     \\
Activation function     & ReLU    \\
Dropout               & 0.2     \\
\bottomrule
\end{tabular}
\caption{Hyperparameters of the NN calibrator}
\label{table:appendix_hyp}
\end{table}

\subsection{Inference}

During inference, we feed context into the parameterized LM, its contextualized representation into the memory index, and its three types of features into the calibrator. We search for top-1K nearest neighbors from 32 nearest cluster centroids using the memory index. The calibrator reweighs the distributions of the parameterized LM and the memory, and we use this calibrated distribution as the final output of our model.

\section{CL baselines}
\label{appendix:cl_baselines}
\textbf{RecAdam} \citep{chen_recall_2020} \ \ As a regularization-based method, RecAdam recalls previously acquired knowledge by retaining the pretraining object through frozen parameters, and it continually learns new information using a multi-task learning object. As the learning process moves forwards, the regularization is annealed to lessen the restriction.

\textbf{Mix-Review} \citep{he2021analyzing} \ \ Assuming that the pretraining corpus is obtainable, Mix-Review uses an empirical decreasing function to adjust the quantity of the pretraining corpus mixed in the continued training data. As the learning process moves forward, the quantity of the pretraining corpus tapers off to 0, resulting in the remaining training process being equivalent to fine-tuning.

Besides established CL baselines, those methods aimed at data efficiency were also considered to be adapted to CL settings, such as Greedy Merging which performs the best in datastore pruning in \citet{he_efficient_2021}. However, while Greedy Merging can be generalized for CL by pruning and merging memory greedily every certain number of steps iteratively, this approach presents disastrous distribution shifts. If we merge new memory into the old, the new information distribution will continuously shift towards the old distribution, finally destroying the performance of new data; vice versa, the old distribution will shift towards the new one, causing the catastrophic forgetting problem. It turns out in Table \ref{tab:cl_baselines} that Greedy Merging undergoes severe catastrophic forgetting in the first 3 months of NC-20H1. 

We leave more dedicated and adapted CL approaches to be explored in the future.

\section{Analysis}

\subsection{Possible learning strategies for MoNIM}
\label{appendix:learn_metric}

In addition to the intuitive cross-entropy loss, there are reasonable methods to measure and control the learning process. We also propose to rely on the intrinsic information content within the memory to assess the necessity of memorization.

\paragraph{Internal information based} Since memorized keys can be projected to vocabulary space to analyze the information hidden in keys, we can calculate the internal distance from key to value, namely the KL-divergence from key-projected token distribution to the golden token distribution, which represents the amount of new information contained in the sample.
\begin{equation}\nonumber
    \hat{M}_{\Theta}(\boldsymbol{\mathrm{x}})=
    \begin{cases}
        1 & \text{if} ~ D_{\mathrm{KL}}(\boldsymbol{y}~||~E\boldsymbol{\mathrm{x_{<t}}}) < \delta \\
        0 & \text{else}
    \end{cases}
\end{equation}

The preliminary results (\cref{table:learning_strategy}) indicate the dominance of cross-entropy loss over KL divergence of internal information. Due to the resource limitation, we stick to the learning strategy using cross-entropy loss in the main experiments throughout the rest of the paper.

\subsection{Performance \textit{VS} Memorization rate}
\label{appendix:memrate_abl}

We have confirmed that MoNIM can achieve performance comparable to \textit{FullMem} with a substantially reduced memorization rate when $\delta=-1.5$. Intuitively, if $\delta$ increases, more cases will be memorized and the performance will likely increase further; on the contrary, if $\delta$ decreases, more cases will be skipped, resulting in less memory but weaker performance. Table \ref{table:abl_threshold} confirms this intuition, demonstrating that we can obtain a trade-off between scalability and performance through the manipulation of $\delta$.

\begin{table}[t]
\centering
\resizebox{\columnwidth}{!}{
\begin{tabular}{l|c|c}
\toprule
Methods     & PPL ($\downarrow$) & MemRate ($\downarrow$) \\
\midrule
Bare GPT-2 small           & 24.1         & 0\%              \\
+MoNIM(loss)  & \textbf{8.6} & \textbf{50\%}    \\
+MoNIM(KL)  & 9.5 & 54\%    \\
\bottomrule
\end{tabular}
}
\caption{Perplexity (PPL) on the test set of NC-20H1 by different learning strategies for MoNIM over NC-20H1.}
\label{table:learning_strategy}
\end{table}

\begin{table}[t]
\centering
\resizebox{\columnwidth}{!}{
\begin{tabular}{l|c|c}
\toprule
Methods     & PPL ($\downarrow$)     & MemRate ($\downarrow$) \\
\midrule
Bare GPT-2 small           & 24.1         & 0\%              \\ \midrule
+\textit{FullMem}      & 9.0          & 100\%            \\ \midrule
+MoNIM ($\delta=-1.0$) & \bf 8.2         & 54\%           \\
+MoNIM ($\delta=-1.5$) & 8.6         & 50\%           \\
+MoNIM ($\delta=-2.0$) & 9.9         & \bf 45\%           \\
\bottomrule
\end{tabular}
}
\caption{MoNIM with different memorization threshold $\delta$.}
\label{table:abl_threshold}
\end{table}

\begin{table}[t]
\centering
\small
\begin{tabular}{c|ccc}
\toprule
Methods     & FullMem     & RandMem      & MoNIM        \\
\midrule
Constant $\lambda$      & 9.0        & 15.0        & 14.3        \\
NN calibrator  & \textbf{8.3} (-0.7) & 12.5 (-2.5) & 8.6 (\textbf{-5.7})  \\
\bottomrule
\end{tabular}
\caption{Perplexity results on the NC-20H1 test data with and without the NN calibrator for FullMem, RandMem, and MoNIM.}
\label{table:abl_cali}
\end{table}

\begin{table}[t]
\centering
\begin{tabular}{l|c}
\toprule
Features               & PPL ($\downarrow$) \\
\midrule
All           & 8.6     \\
-Density features      & 12.0     \\
-Distribution features & 9.9      \\
-Lexical features      & 8.9      \\
\bottomrule
\end{tabular}
\caption{The ablation study of features in the NN calibrator.}
\label{table:abl_features}
\end{table}

\begin{figure}[t]
\centering
\includegraphics[width=\columnwidth]{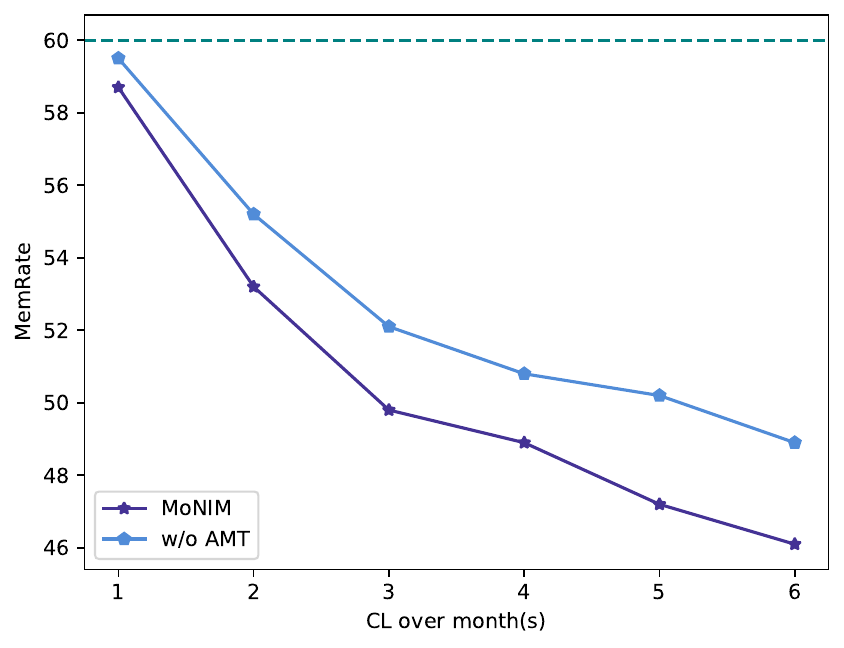}
\caption{MoNIM's memorization rate with CL over months.}
\label{fig:lm_grad}
\end{figure}

\subsection{NN calibrator}
\label{appendix:cali_abl}

The NN calibrator plays an important role in calibrating the semiparametric LM's prediction by adapting $\lambda$ at test time. Although such an adaptive method has been proven universally effective in semiparametric LMs by previous work, we reveal in Table \ref{table:abl_cali} that it benefits MoNIM most, significantly more than \textit{RandMem} and conventional \textit{FullMem}, which only introduces marginal improvement as in previous work, demonstrating that calibration is more compatible with MoNIM.

We ablate the features of the NN calibrator to study their effects on the results. According to Table \ref{table:abl_features}, all our proposed features contribute positively to the calibrator, among which the density information, especially the $L^2$ distance, is the most important one because it can directly reflect if the non-parametric memory has much relevant information given a test case, providing the most straightforward evidence to the calibrator. 

\begin{table}[t]
\centering
\small
\begin{tabular}{l|ccc}
\toprule
Methods (MemRate)     & 0-shot   & 1-shot   & 2-shot  \\ 
\midrule
Bare GPT-2 large (0\%)           & 29.8     & 31.5     & 32.5    \\
+\textit{FullMem} (100\%)      & 36.3     & \textbf{36.9}     & 37.2    \\
+\textit{RandMem} (60\%)      & 30.7     & 32.1     & 33.0     \\ \midrule
+MoNIM (40\%) & \textbf{36.3} & 36.7 & \textbf{37.7}    \\
\bottomrule
\end{tabular}
\caption{In-context learning accuracy of the GPT-2 large with CL over NC-20H1 on the \textsc{RealTime QA} in 0-, 1- and 2-shot learning.}
\label{table:in_context_qa}
\end{table}

\subsection{Adaptive memorization threshold (AMT)}
\label{appendix:amt}

We test MoNIM's MemRate with/without AMT, since it serves as a measure of AMT's performance in handling memory overfitting. It reveals in \cref{fig:lm_grad} that after adding AMT the downward trend is largely enhanced, while the performance is nearly untouched despite the reduced memory. We claim that the overfitting causes the predictions to favor the neighboring overconfident wrong answers, damaging the generalization of new data. Thus, the alleviation of overfitting is helpful with respect to both performance and scalability.

\subsection{In-context learning}
\label{appendix:icl}

We test if the in-context learning capability of a language model is affected by CL through MoNIM. We present the in-context learning result in the \textsc{RealTime QA} benchmark that the memory can benefit in Table \ref{table:in_context_qa}, showing that MoNIM is not in conflict with in-context learning and that the LM can still benefit from more examples shown in the context.

Also, we evaluate MoNIM's effect on in-context learning for general tasks collected by ~\citet{shi2022nearest} that cannot benefit from its memory. According to Table \ref{table:in_context_multitask}, despite no improvement observed, MoNIM does not affect the results in these tasks, demonstrating its robustness.

\begin{table*}[t]
\centering
\begin{tabular}{l|cccccccc}
\toprule
Methods (MemRate)  & RTE  &  CB  &  RT  & SST-2 & CR  & MR   & HYP  \\
\midrule
Bare GPT-2 large (0\%)        & 53.1 & 39.3 & 49.5 & 51.4 & 50.5 & 50.8 & 60.0 \\
\textit{FullMem} (100\%)   & 50.3 & 35.5 & 49.0 & 49.8 & 48.6 & 47.5 & 60.0 \\
\textit{FullMem} (100\%) w/ NN calibrator 
          & 52.8 & 41.1 & 49.5 & 51.8 & 50.8 & 50.9 & 60.0 \\ \midrule
MoNIM (40\%)     & 53.1 & 41.1 & 49.5 & 51.5 & 50.8 & 50.0 & 60.0 \\
\bottomrule
\end{tabular}
\caption{0-shot learning accuracy of the GPT-2 large with CL over NC-20H1 on general NLP tasks.}
\label{table:in_context_multitask}
\end{table*}

\subsection{Time efficiency}
\label{appendix:efficiency}

We analyze the time consumption of MoNIM with vanilla kNN-LMs.

\begin{itemize}
    \item Index Building: In this process, the whole computational overhead is equal to conducting a full forward pass over the training data to extract representations as keys, which is the same as a vanilla kNN-LM.
    \item Retrieval Process: We have measured that MoNIM’s inference time is approximately $0.9$–$1.1\mathtt{x}$ that of vanilla kNNs. For time-sensitive tasks, we have also explored a simple modification to accelerate inference without significantly hurting performance. We introduce a confidence threshold $\theta$ after the calibrator of $\lambda$ in \cref{sec:ada}, so that if $\lambda < \theta$, we simply skip the retrieval step and rely solely on the LM output. This is intuitive since a low $\lambda$ indicates that memory contributes little useful information. In our experiments with GPT-2 small, when $\theta = 0.3$, the inference latency is reduced to $0.8\mathtt{x}$ that of vanilla kNNs, while the PPL in \cref{table:lm_compare_strategy} increases slightly from 8.6 to 8.9, while \textit{FullMem} PPL is 9.0.

\end{itemize}

\end{document}